\documentclass[conference]{IEEEtran}
\IEEEoverridecommandlockouts
\usepackage{cite}
\usepackage{amsmath,amssymb,amsfonts}
\usepackage{algorithmic}
\usepackage{graphicx}
\usepackage{textcomp}
\usepackage{xcolor}
\usepackage{cleveref}
\usepackage{booktabs}
\usepackage{url}

\def\BibTeX{{\rm B\kern-.05em{\sc i\kern-.025em b}\kern-.08em
    T\kern-.1667em\lower.7ex\hbox{E}\kern-.125emX}}

\makeatletter 
\newcommand{\linebreakand}{%
  \end{@IEEEauthorhalign}
  \hfill\mbox{}\par
  \mbox{}\hfill\begin{@IEEEauthorhalign}
}
\makeatother 

\usepackage{fancyhdr}

\fancypagestyle{withfooter}{

\fancyfoot[C]{\footnotesize Accepted to the Challenges and Opportunities of Neuromorphic Field Robotics and Automation IEEE ICRA Workshop - 2026}
}
\begin{document}

\title{Neuromorphic Monocular Depth Estimation with Uncertainty Modeling
}

\author{ \IEEEauthorblockN{1\textsuperscript{st} Viktor Bergkvist}
\IEEEauthorblockA{\textit{Swedish Defence Research Agency} \\
Linköping, Sweden \\
viktor.bergkvist@foi.se}
\and
\IEEEauthorblockN{2\textsuperscript{nd} Felix Rydell}
\IEEEauthorblockA{
\textit{Swedish Defence Research Agency}\\
Kista, Sweden  \\
felix.rydell@foi.se}
\and
\IEEEauthorblockN{3\textsuperscript{rd} Per-Erik Forssén}
\IEEEauthorblockA{
\textit{Linköping University}\\
Linköping, Sweden \\
per-erik.forssen@liu.se}
\and
\linebreakand
\IEEEauthorblockN{4\textsuperscript{th} David Gustafsson}
\IEEEauthorblockA{\textit{Swedish Defence Research Agency}\\
Linköping, Sweden \\
david.gustafsson@foi.se}
\and
\IEEEauthorblockN{5\textsuperscript{th} Johan Rideg}
\IEEEauthorblockA{\textit{Swedish Defence Research Agency}\\
Linköping, Sweden \\
johan.rideg@foi.se}}

\maketitle


\begin{abstract}
Event cameras offer distinct advantages over conventional frame-based sensors, including microsecond-level temporal resolution, high dynamic range, and low bandwidth. In this paper, we predict per-pixel depth distributions from monocular event streams using deep neural networks. We estimate uncertainty using Gaussian, log-normal, and evidential learning frameworks. We compare six event representations: spatio-temporal voxel grids with 1, 5, 10, and 20 temporal bins, the Compact Spatio-Temporal Representation (CSTR), and Time-Ordered Recent Event (TORE) volumes. Our U-Net-based models are trained on synthetic data and then fine-tuned on real sequences. We evaluate performance using absolute relative error, root mean squared error, and the area under the sparsification error. Quantitative results show that the representations perform similarly, while 10 bin log-normal and 5 bin evidential learning perform best across metrics. Our experiments demonstrate that uncertainty estimation can be successfully integrated into event-based monocular depth estimation, and be used to indicate pixels with reliable depth.
\end{abstract}



\section{Introduction}
Camera based depth estimation is a fundamental computer vision problem
with many practical applications. It is an enabling technique for localization, path
planning, and collision avoidance, which in turn are used for
autonomous systems, including self-driving vehicles and robotics, and
augmented reality platforms. However, in many real-world scenarios, conventional frame-based cameras struggle under challenging lighting conditions and highly dynamic environments, where motion blur and limited dynamic range can degrade depth estimation performance.

Event cameras provide a promising alternative to conventional frame-based sensors for depth estimation. These devices are built on a relatively recent imaging technology inspired by biological vision systems~\cite{Event-survey}. Unlike traditional cameras, event cameras operate asynchronously: each pixel independently monitors changes in brightness and triggers an \textit{event} whenever a significant intensity variation is detected. This sensing principle offers several benefits. For instance, event cameras provide microsecond-level temporal resolution, enabling the capture of fast motion without inducing blur. Because each pixel acts independently, event cameras achieve very low latency, without the need to wait for a global shutter or exposure cycle. Furthermore, they exhibit an exceptionally high dynamic range, which allows them to operate more effectively across challenging lighting conditions.

Monocular depth estimation infers scene depth collected from a single camera, which is particularly relevant for small surveillance drones, where mounting multiple cameras sufficiently far apart to provide a meaningful stereo baseline is often impractical.

To the best of our knowledge, we are the first to do monocular depth estimation with uncertainty using event cameras. We train neural networks to predict depth by optimizing one of three per-pixel probabilistic models: Gaussian, log-normal, and evidential distributions. These approaches are compared against a model trained on grayscale images and a baseline depth estimation model that predicts depth without uncertainty.

Due to the asynchronous and sparse nature of event data, methods developed for frame-based cameras cannot be applied directly to event data. Instead, one typically converts the raw event stream into an {\it event representation} that makes the input compatible with existing deep learning architectures. We consider and compare six event representations: spatio-temporal voxel grids with varying temporal bin sizes~\cite{zhu2018unsupervisedeventbasedlearningoptical,Hidalgo20threedv}, the compact spatio-temporal representation (CSTR)~\cite{cstr}, and time-ordered recent event volumes (TORE)~\cite{TORE}.

Supervised depth prediction requires a large amount of annotated event data. Obtaining such ground-truth data at scale is often impractical~\cite{DepthAnything}. One strategy to circumvent limited data is to pretrain on simulated data and fine-tune on real data. In our work, we leverage the BlinkVision synthetic indoor video dataset by Li et al., which provides pixel-perfect ground-truth depth~\cite{li2024blinkvisionbenchmarkopticalflow}. All scenes are rendered entirely in Blender\footnote{\url{https://www.blender.org/}}. Event data for these videos is then generated using the physics-based DVS-Voltmeter simulator, which models the circuitry of a real dynamic vision sensor and is chosen for its realistic noise characteristics~\cite{DVS-Voltmeter}. We fine-tune our depth estimation models on a real-world dataset to expose them to realistic noise and domain shift~\cite{MVSEC}. We consider the indoor setting in this paper, due to the convenience of finite depths.

\subsection*{Contributions} Our main contributions are listed as follows: 

\begin{itemize}
    \item \textbf{Uncertainty estimation:} We compare three different uncertainty models for depth estimation: Gaussian, log-normal, and an evidential learning approach. These exhibit similar performance across the board, although the results depend on the choice of event representation.
    
    \item \textbf{Event representations:} We compare six different event representations for monocular depth estimation: spatio-temporal voxel (1, 5, 10, 20 bins), CSTR, and TORE. Our experiments suggest that each representation is a valid choice for depth estimation, although performance varies depending on the uncertainty model.  
\end{itemize}

\subsection*{Organization} This paper is structured as follows. In \Cref{sec: rel work}, we explore related work. In \Cref{sec:method}, we define the different event representations and the model architecture we use, as well as training details. In \Cref{sec:exp}, we explain the experiments we conduct. In \Cref{sec:results}, we present the qualitative and quantitative results. Additional results are provided in the Supplementary Material. Finally, in \Cref{sec:discussions_and_future_work}, we summarize and conclude the paper.


\section{Related Work}\label{sec: rel work}

\subsection{Geometric-based multiview depth estimation from events} Early approaches to depth estimation with event cameras rely on adapting the asynchronous event data for classical geometric methods, i.e., feature matching, pose estimation, and triangulation. These typically produce sparse or semi-dense depth measurements rather than full dense depth maps~\cite{Hidalgo20threedv}. To match events, some authors use standard stereo vision techniques applied to synthetic frames created by accumulating events over time~\cite{Event-Based-Stereo-Vision-with-Bio-Inspired-Silicon-Retina-Imagers}, while others leverage the simultaneous and temporal correlation of events detected across the different sensors~\cite{Ontheuseoforientationfiltersfor3Dreconstructioninevent-drivenstereovision,AsynchronousEvent-BasedBinocularStereoMatching}. Rebecq et. al.~\cite{Rebecq18ijcv} introduced an event-based multi-view stereo algorithm that accumulates events on a depth volume, assuming the cameras' motions are known. 

Another strategy is to compute optical flow and convert it to depth via motion equations. Gallego et al.~\cite{CMapplications,CMfocus} proposed a contrast maximization framework, where a hypothesized optical flow is evaluated by how sharply it warps events into an image. Contrast maximization has, however, exhibited a tendency to overfit~\cite{zhu2018unsupervisedeventbasedlearningoptical}. Shiba et al.~\cite{Shiba22eccv} tackle this problem by partitioning the image into tiles and introducing a multi‑reference focus loss that averages edge sharpness over three timestamps, ensuring that motion‑compensated images remain sharp.

Inspired by DTAM~\cite{newcombe2011dtam}, Zhou et al.~\cite{zhou2018semi} proposed to maximize photometric consistency by optimizing a certain energy function in a stereo setting. Their algorithm produces a semi-dense reconstruction by exploiting spatio-temporal consistency of events triggered across both image planes. 

Cai and Bideau~\cite{ActiveEventAlignment} propose a region-based monocular depth estimation using an active vision approach, mimicking biological gaze stabilization.

\subsection{Learning-based monocular depth estimation from events} Deep learning has enabled a shift from sparse, geometric methods toward dense monocular depth prediction using event data. Broadly speaking, current approaches fall into three categories: fully supervised networks trained on paired event- and depth data, multimodal fusion architectures that combine events with for example IMU data, and self-supervised methods that exploit physical consistency in the event stream.

In~\cite{zhu2018unsupervisedeventbasedlearningoptical}, Zhu et al. developed an unsupervised framework that simultaneously predicts optical flow, ego-motion, and sparse depth from the events. Their method introduced a discretized event volume (voxel grid) representation that captures the spatio-temporal distribution of events.

Hidalgo-Carrio, Gehrig, and Scaramuzza introduced E2Depth, a supervised approach using recurrent encoder-decoder model (based on the U-Net architecture~\cite{UNET}) with ConvLSTM layers that process a spatio‑temporal voxel grid of events~\cite{Hidalgo20threedv} that was trained on a large synthetic dataset. We choose this model as the baseline for our paper, because it is the only method that uses depth supervision and is purely event based. Another supervised method by Gehrig et al.~\cite{RAMNet} used an asynchronous recurrent network that integrated both event streams (as voxel grids) and regular intensity frames. 

Departing from purely data-driven approaches, Meng et al. utilize sensor ego-motion~\cite{meng2024learning}, constructing a cost volume based on the sharpness of Image of Warped Events (IWE) across various depth hypotheses. 

Bartolomei et al.~\cite{bartolomei2025depth} make use of a Vision Foundation Model to create synthetic depth data and to train an event-based student model with a recurrent transformers-based architecture. Like us they also fine-tune on MVSEC~\cite{MVSEC}.


\section{Methodology}\label{sec:method}First, we briefly describe the different event representations used in this paper. Second, we present the model architecture we use to predict depth and uncertainty.


\subsection{Event representations}\label{subsec:event_rep}

Consider an event camera image plane of size $H\times W$. An event $e_k=(\mathbf{u}_k,t_k,p_k)$ consists of a pixel position $\mathbf{u}_k\in H\times W$, a timestamp $t_k\in\mathbb R$, and a polarity $p_k\in \{-1,+1\}$. Suppose that $I(\mathbf{u}_k, t_k)$ denotes the light intensity at pixel $\mathbf u_k$ at time $t_k$, and let $\Delta t$ denote the time since the last event in that pixel. In a noise‐free scenario, the event is triggered if the brightness increment since its last event
\begin{equation}
\Delta \log I(\mathbf{u}_k,t_k):
=
\log I(\mathbf{u}_k,t_k)
-\log I\bigl(\mathbf{u}_k,t_k-\Delta t_k\bigr),
\label{eq:event1}
\end{equation}
reaches a signed threshold $p_kC$, for some $C>0$, meaning that $\Delta \log I(\mathbf{u}_k,t_k)=p_kC$~\cite{Event-survey}. The polarity $p_k$ indicates whether the brightness increased or decreased. An event batch $\mathcal{E}=\{(\mathbf{u}_i,t_i,p_i)\}$ is a collection of events over a time-window $\Delta T$. Below we present three event representations, which provide useful structure, but inevitably result in information loss. We use the notation $\mathbf{1}_{x=y}$ to be 1 precisely when $x=y$ holds and 0 otherwise. 

\textbf{Spatio-temporal voxel grids.} The idea of voxel grids is to divide a time-window into bins and simply count events in those, normalizing with respect to time~\cite{zhu2018unsupervisedeventbasedlearningoptical,Hidalgo20threedv}. In this direction, partition the time-window $\Delta T$ into equally sized temporal bins indexed by $b\in \{0,1,\ldots, B-1\}$. For a timestamp $t_i$, the bin function $k_b(t_i):=\max(0, 1 - |b - t^*_i|)$, where $t_i^* = \frac{B - 1}{\Delta T}(t_i - t_0)$ is the normalized time, is $0$ for any $t$ that is outside the interval $[b,b+1]$. Inside a bin, $k_b(t_i)$ tells us how far away $t_i$ is from the bin end time. Using this bin function, we define the spatio-temporal voxel grid as the $B\times H\times W$-tensor 
\begin{equation}
E(b, \mathbf{u}) = \sum_{i=1}^{N} p_i \mathbf{1}_{\mathbf{u} = \mathbf{u}_i}k_b(t_i).
\label{eq:voxel_grid}
\end{equation} 

The use of spatio-temporal voxel grids have two primary challenges. First, generating spatio-temporal voxel grids can be computationally demanding. Second, using voxel grids may lead to high memory requirements due to high input dimensionality~\cite{Rebecq18ijcv,yin2024exploring}.

\textbf{Compact spatio-temporal representation (CSTR).} In constrast to voxel grids, CSTR does not bin events temporally, but more clearly distinguishes between negative and positive polarities. Consider the per-polarity event count $C(\mathbf{u},p)$ and the normalized event counts $\hat C(\mathbf{u})$:
\begin{align}
\begin{aligned}
C(\mathbf{u},p)&:
=
\sum_{i=1}^N\mathbf{1}_{\mathbf{u}_i=\mathbf{u}} \mathbf{1}_{p_i=p},\\ 
\hat C(\mathbf{u})
&:=
\frac{C(\mathbf{u},-1) + C(\mathbf{u},+1)}{\max_{\mathbf{u}}\{C(\mathbf{u},-1) + C(\mathbf{u},+1)\}}.
\end{aligned}
\label{eq:cstr_per_prixel}
\end{align}
We further define the sum of normalized timestamps $S(\mathbf{u},p)$ and the pixel-wise mean normalized timestamp $\bar T_s(\mathbf{u},p)$ for each polarity:
\begin{align}
\begin{aligned}
S(\mathbf{u},p)&
:=
\sum_{i=1}^N\mathbf{1}_{\mathbf{u}_i=\mathbf{u}} \mathbf{1}_{p_i=p}
\frac{t_i - t_0}{\Delta T}
,\\ \bar T_s(\mathbf{u},p)&:
=
\begin{cases}
\dfrac{S(\mathbf{u},p)}{C(\mathbf{u},p)}, & C(\mathbf{u},p)\neq 0,\\[6pt]
0, & \text{otherwise}.
\end{cases}    
\end{aligned}
\label{eq:cstr_meants}
\end{align}
Finally, the CSTR is constructed by concatenating the two polarity‐specific mean‐timestamp channels with the normalized event‐count channel~\cite{cstr}:
\begin{equation}
\mathrm{CSTR}(\mathbf{u})
:=
\begin{bmatrix}
    \bar T_s(\mathbf{u},+1) & & \hat C(\mathbf{u}) & & \bar T_s(\mathbf{u},-1)
\end{bmatrix},
\label{eq:cstr}
\end{equation}
yielding a compact $3\times H\times W$ tensor. 

Flickering light sources and hot pixels can heavily affect the normalized event-count channel, still, CSTR preserves temporal context under high event overlap and remains directly compatible with off‐the‐shelf three‐channel architectures.

\textbf{Time-ordered recent event (TORE) volumes.} \label{the:tore} Time surfaces are $H\times W$ matrices where each entry encodes the time of the latest event at that location. Time-ordered recent event (TORE) volumes generalize these by storing the times of the $K$ latest events, and making a distinction between polarity~\cite{TORE}. In more detail, for a fixed polarity $p$ and pixel $\mathbf{u}$, let $t_k$ be time of the $k$-th most recent event at $\mathbf{u}$ of polarity $p$ at time $t$. The TORE volume is then defined as the $2\times K\times H\times W$ tensor 
\begin{align}
    \mathrm{TORE}(p,k,\mathbf{u}):= \mathrm{log}(t-t_k + 1).
\end{align}
To manage noise and computational efficiency, these values are clipped from below by $\mathrm{log}(\tau')$ and from above by $\mathrm{log}(\tau)$, for some $\tau,\tau'$.

\subsection{Neural network architecture}\label{subsec:models}
\begin{figure*}[t]
    \centering
    \includegraphics[width=0.75\textwidth]{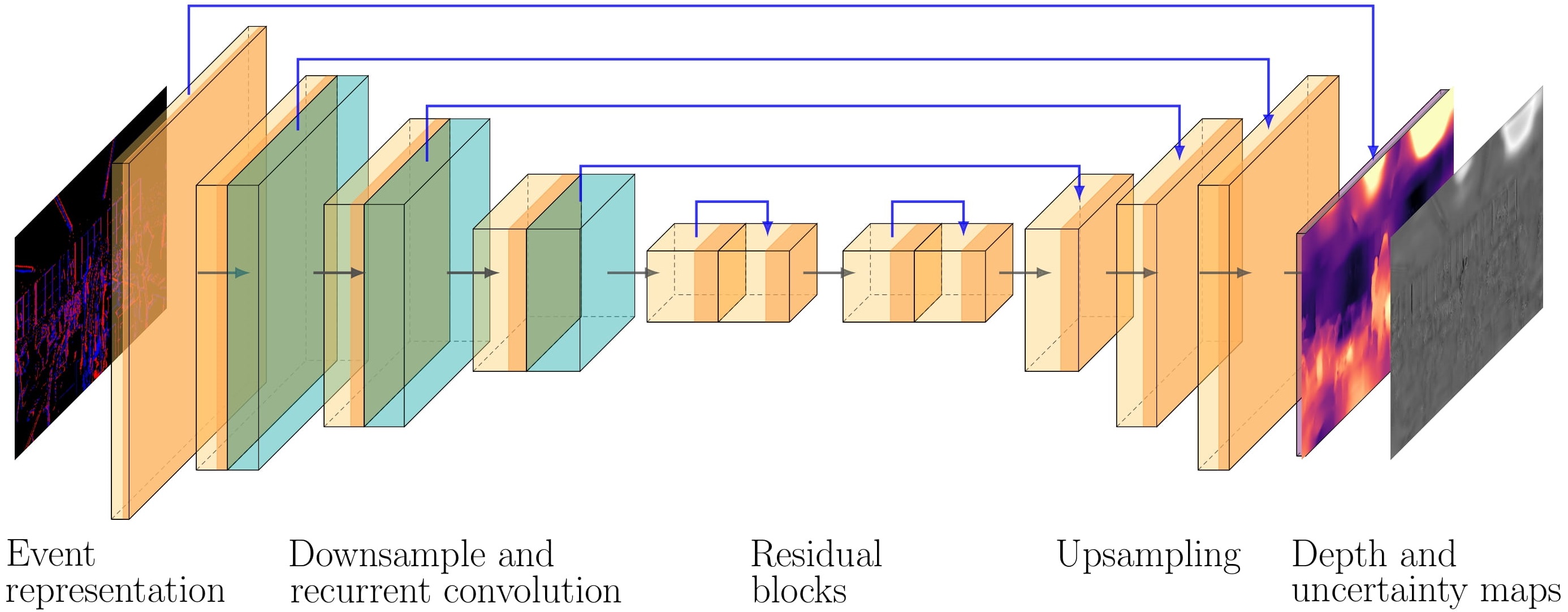}
    \caption{Overview of the network architecture used for predicting depth and uncertainty. The blue arrows denote skip connections.}
    \label{fig:model_architecture}
\end{figure*}

The depth prediction model we use is a recurrent encoder-decoder framework inspired by~\cite{Hidalgo20threedv}, structured similarly to U-Net~\cite{UNET}. The input event data representation first passes through a convolutional head layer, producing an initial feature map with 32 channels. Following that, the architecture consists of three recurrent convolutional encoder layers, each composed of a downsampling convolution with a kernel size of 5 and stride of 2, followed by a ConvLSTM module with a kernel size of 3. Each encoder layer doubles the number of feature channels from the previous layer, resulting in a final encoded representation of 256 channels. See \Cref{fig:model_architecture}.

Following the encoder layers, two residual blocks are applied, each using convolutional operations with a kernel size of 3 and including summation over skip connections to facilitate gradient flow and feature refinement. Afterward, the architecture includes three decoder layers, each performing bilinear upsampling followed by convolutional refinement with a kernel size of 5. Symmetric skip connections with summation are used between corresponding encoder and decoder layers to enhance the preservation of spatial information.

Finally, the output is produced by a convolution whose number of channels depends on the uncertainty formulation and a kernel size of 1. For Gaussian and log-normal models the network predicts two channels corresponding to the depth estimate and an uncertainty parameter, while the evidential model predicts four channels representing the parameters of a Normal–Inverse-Gamma distribution. The network is unrolled for 20 time steps to capture temporal coherence from the asynchronous event stream. ReLU activations are used throughout, except for the final prediction layer.

The networks predict depth directly in metric units, and the ground-truth depths from the dataset are used without normalization during training. This simplifies the interpretation of both predicted depths and uncertainty estimates. In contrast, the network in ~\cite{Hidalgo20threedv} predicts a normalized log-depth map $\hat{D}_k \in [0,1]^{W \times H}$, which is converted to metric depth through the following mapping:
\begin{equation}
\hat{D}_{m,k} = D_{\max}e^{(-\alpha (1-\hat{D}_k))}.
\label{eq:e2depth_mapping}
\end{equation} 
In our experiments, we retain this formulation for the E2Depth baseline, using $D_{\max}=35$ and $\alpha=8$, which were chosen to better suit the indoor depth range considered in this work.


\subsection{Uncertainty prediction and loss functions} \label{subsec:loss}

Three different depth distribution prediction networks are trained by optimizing the losses below. In our experiments, they are compared to the E2Depth~\cite{Hidalgo20threedv} approach, which minimizes squared depth errors, the sum of depth errors, and a multi-scale gradient-matching loss.

\textbf{Gaussian depth estimation.} The Gaussian depth estimation predicts mean $\mu_k(\theta)$ and standard deviation $\sigma_k(\theta)>0$ (we apply softplus to enforce positivity throughout), with objective
\begin{align}
    f_{\mathrm{GD}}(\theta,k) :=\frac{1}{\sigma_k(\theta)\sqrt{2\pi}}
e^{-\tfrac{(y_k-\mu_k(\theta))^2}{2\sigma_k(\theta)^2}}.
\end{align}
Depth is estimated as $\mu_k(\theta)$ and variance as $\sigma(\theta)^2$.

\textbf{Log-normal depth estimation.} Depth cannot be negative, therefore log-normal depth estimation network is a natural alternative. It outputs $\mu_k(\theta)$ and scale $\sigma_k(\theta)> 0$, with objective
\begin{align}
    f_{\mathrm{LD}}(\theta,k) :=\frac{1}{y_k\sigma_k(\theta)\sqrt{2\pi}}
e^{-\tfrac{(\log y_k-\mu_k(\theta))^2}{2\sigma_k(\theta)^2}}.
\end{align}
The depth is estimated as the mean $e^{\mu_k(\theta) + \sigma_k(\theta)^2/2}$ of the log-normal distribution, and the variance is $(e^{\sigma_k(\theta)^2} - 1)e^{2\mu_k(\theta) + \sigma_k(\theta)^2}$.

\textbf{Evidential depth estimation.} There are essentially two types of errors: aleatoric (inherent in data) and epistemic (due to model). Evidential deep learning (EDL) constitutes an attempt to separate these two uncertainties~\cite{sensoy2018evidential}. Our evidential approach estimates four parameters: location $m(\theta)$, evidence $\kappa(\theta)> 0$, confidence about variance $\alpha(\theta)>1$, and level of noise $\beta(\theta)>0$. In this case, the objective is 
\begin{align}
    f_{\mathrm{ED}}(\theta,k) := f_{\mathrm{Student-t}}(y_k \:|\: \mathrm{location},\:\mathrm{scale},\:\nu_{\mathrm{df}}),
\end{align}
where $\mathrm{location}=m$, $\mathrm{scale}=\sqrt{\frac{\beta(1+\kappa)}{\kappa\alpha}}$, and $\nu_{\mathrm{df}}= 2\alpha$. Depth is estimated as the mean $m$, and the variance is estimated as $\frac{\beta}{\kappa(\alpha - 1)}$. We interpret $\frac{\beta}{\alpha - 1}$ as the aleatoric error and $\frac{\beta}{\kappa(\alpha - 1)}$ as the epistemic error. Evidential deep learning has previously been used for monocular depth estimation using RGB frames~\cite{amini2020deep}.

$ $

For the probabilistic objectives
$f_{\mathrm{GD}}$, $f_{\mathrm{LD}}$, and $f_{\mathrm{ED}}$, we minimize the negative-loglikelihood during training:
\begin{equation}
\mathcal{L} = - \frac{1}{N} \sum_{k=1}^N \log f(\theta,k).
\end{equation}
In the evidential learning case, we add a term $\lambda |y_k-m|(2\kappa +\alpha)$, for $\lambda = 0.25$, which heavily penalizes the objective if $ |y_k-m|$ is big, and guides the network to output small $\kappa$ when predictions are poor or data is scarce.


\section{Experiments}\label{sec:exp}

We predict per-pixel depth distributions from event data by training the networks described in \Cref{subsec:models} in two stages. First we train on a synthetic dataset, and second we fine-tune on real data. We use the loss functions described in Section \ref{subsec:loss}. For the log-normal loss, we consider all event representations to be able to compare and contrast the performance of the different representations. For the Gaussian and evidential approaches, we restrict to 5 bin and CSTR, while for E2Depth, we restrict to 5 bin only. We train the E2Depth baseline using the same synthetic pretraining and MVSEC fine-tuning protocol as the proposed models, rather than using an external pretrained checkpoint. Since the real dataset contains grayscale images, we also train a model on these using the log-normal loss. The models are evaluated using two error measurements for depth prediction and two error measurements for the uncertainty modelling. We describe datasets, data preprocessing, and evaluation metrics below.

All experiments were implemented in PyTorch and were conducted on a workstation equipped with an Intel Core Ultra 9 285K CPU, 96 GB of RAM, and an NVIDIA RTX 5090 GPU. Training was run for 200 epochs with a batch size of 20 and with the Adam optimizer~\cite{kingma2015adam}. The models are first trained on synthetic data with learning rate $10^{-4}$. During fine-tuning on real data the learning rate is reduced to $10^{-5}$ and weight decay of $10^{-5}$ is applied.

\subsection{Datasets}\label{subsec:data}

\textbf{BlinkVision.} BlinkVision provides photorealistic simulated renderings~\cite{li2024blinkvisionbenchmarkopticalflow}. It contains 24 indoor sequences of length ranging from 10 to 25 seconds. We use 20 sequences for training and 4 for validation. For every rendered viewpoint the dataset has an HDR RGB frame, a metric depth map as well as several other modalities (normals, object masks, etc.). Each sequence is internally sampled and interpolated to a higher frame rate for the event generation. Event streams for the sequences are generated with the physics-based DVS-Voltmeter simulator, chosen for its realistic noise model~\cite{DVS-Voltmeter}. 

In some frames, we have observed that isolated pixels or groups of pixels have falsely high values far beyond the scene range, while at geometric discontinuities depth is occasionally reported as exactly zero. The choice of parameter \(D_{\max}=35\) clips all synthetic depths, helps mitigate the former issue. In addition, we mask out zero‐valued pixels during both training and evaluation.

\textbf{MVSEC indoor.} The indoor‐flying sequences of the Multi Vehicle Stereo Event Camera (MVSEC) dataset provide real-world event and depth data captured in a motion-capture arena~\cite{MVSEC}. The scenes typically depict a room with objects such as barrels and chairs. The sequences are captured by a hexacopter carrying a stereo DAVIS-346B rig (event cameras) and a Velodyne Puck LITE LiDAR ($360^\circ$ horizontal FOV, $30^\circ$ vertical FOV, 20 Hz) mounted overhead. Ground-truth 6 DoF pose is measured at 100 Hz by a Vicon motion-capture system. There are four sequences of length ranging from about 20 to about 94 seconds. 2 are used for training, 1 for validation, and 1 for test (indoor flying 1). MVSEC contains grayscale images, which we also train on to compare results with event data.

In contrast to the synthetic data, LiDAR returns are sparse with a limited vertical field of view.

\subsection{Data preprocessing}
Samples consist of non-overlapping windows of events spanning $\Delta T = 50$ milliseconds, converted into the different representations described in Section \ref{subsec:data}. This means that we use event data from such intervals to estimate depth at the end of the interval. Spatio-temporal voxel grids with different temporal resolutions (1, 5, 10, 20 bins) are generated to explore the impact of temporal resolution on depth prediction performance. The non-zero values in the voxel grids are normalized to have zero mean and unit variance. For the TORE volumes, we fix $K=3$ (as in~\cite{TORE-stereo}). We consider $\tau = 5 \times 10^4$ microseconds (this choice differs from~\cite{TORE}, but suits our setting better) and we set $\tau' = 150$. The  volume of shape $2\times K\times H\times W$ is collapsed into $2K\times H\times W$ by concatenating the polarity and temporal dimensions.

The depth maps and the different event representations from the BlinkVision dataset were spatially downsampled from $960\times540$ to $480\times270$, enabling training in a reasonable time. Depth maps were downsampled using nearest-neighbour interpolation to prevent the interpolation of new depth values, thus retaining discrete depth levels. TORE representations specifically utilized min pooling for downsampling. This choice was intentional to retain the most recent event timestamp in each layer, thereby preserving temporal precision and preventing thin structures or event lines from becoming overly smoothed. Voxel grids and CSTR representations, employed bilinear interpolation for downsampling.

Random horizontal flips to both synthetic and real samples are applied to improve generalization.


\subsection{Metrics for evaluation}\label{ss: metrics} In the Gaussian depth estimation case, assume that $ y_k$ for $k=1,\ldots,N$ are per-pixel real depth and $\hat y_k$ are estimated depths with standard deviation $\sigma_k$. We compute absolute relative errors (AbsRels) and root mean squared errors (RMSEs) as
\begin{align}
\begin{aligned}
    \text{AbsRel} &:= \frac{1}{N}\sum_{k=1}^{N}\frac{|y_k - \hat{y}_k|}{y_k},\\ \text{RMSE} &:= \sqrt{\frac{1}{N}\sum_{k=1}^{N}(y_k - \hat{y}_k)^2}.
\end{aligned}
\end{align}

We also consider the Area Under the Sparsification Error (AUSE), which quantifies how well a model's predicted uncertainty correlates with its prediction errors~\cite{Xiliang}. Let $\mathrm{Err}$ denote the mean absolute error (MAE) $\frac{1}{N}\sum_{k=1}^N |y_k-\hat y_k|$. Let $\alpha\in [0,1]$ and define $\mathrm{Err}_{\mathrm{pred}}(\alpha)$ be the MAE of the \( (1 - \alpha) \)-fraction of the dataset with the lowest predicted uncertainty (variance), and $\mathrm{Err}_{\mathrm{oracle}}(\alpha)$ be the MAE of the corresponding fraction the lowest prediction errors. In symbols, the AUSE is:
\begin{equation}
\text{AUSE}= \int_0^1 \left( \frac{\mathrm{Err}_{\mathrm{pred}}(\alpha)- \mathrm{Err}_{\mathrm{oracle}}(\alpha)}{\mathrm{Err}}\right) d\alpha.
\end{equation}
While the AUSE is effective for comparing the trustworthiness of model outputs, it has limitations. It lacks a clear upper bound and its interpretability is relative rather than absolute~\cite{Xiliang}. In the EDL case, $\mathrm{Err}_{\mathrm{pred}}(\alpha)$ is computed with respect to the fraction with the lowest predicted epistemic error.


\section{Results}\label{sec:results} 
The quantitative results for the different uncertainty models and event representations are summarized in \Cref{tab:my_depth_results}. The table reports error metrics for various combinations of uncertainty models and event representations. For each configuration, we evaluate performance over three pixel subsets: (1) all pixels in the image; (2) only pixels that generated events within the considered time window, denoted by (M), which account for approximately 10 \% of all pixels; and (3) the 10 \% of pixels with the lowest predicted uncertainty, denoted by (C), selected to match the proportion of event-triggered pixels in (M).

Among the log-normal variants, the 10-bin voxel grid achieves the best overall performance. The degradation observed with 20 bins could stem from the increased representational complexity of the input, which could make optimization more challenging within the same training schedule. The CSTR and TORE representations yield comparable results, although TORE shows a slight advantage on the uncertainty-conditioned subset (C).

For the Gaussian models, performance appears largely independent of the event representation. However, CSTR yields better results on the (C) metrics compared to the 5-bin voxel grid. This behavior is not observed for the log-normal or EDL formulations. The EDL 5-bin configuration achieves the lowest overall error, although it does not obtain the best AUSE score. In contrast, EDL with CSTR performs comparably to the log-normal model with 20 bins and represents one of the weaker configurations.

These results suggest practical trade-offs rather than a single best event representation. Voxel grids provide a simple and effective baseline; in our experiments, intermediate temporal resolution was preferable, with 10 bins performing best for the log-normal model, whereas 20 bins did not improve accuracy despite retaining finer temporal structure. Since no representation consistently dominates across uncertainty models and metrics, the final choice may depend on system-level constraints such as memory use, inference speed, and the complexity of constructing the representation in the target application.

The event-masked metrics (M) provide a coarse test of whether pixels with recent event activity are systematically easier or harder to predict. The gap between dense evaluation and event-masked evaluation is relatively small across methods, indicating that predictions on event-generating pixels are broadly representative of predictions over the full valid image. Thus the presence of events at a pixel does not by itself strongly explain prediction quality. This does not rule out finer-grained dependencies such as local event rate or polarity balance.

The baseline E2Depth model achieves the best overall performance, with a 21\% lower RMSE compared to the log-normal 10-bin configuration. This gap should be interpreted with care: E2Depth predicts depth and includes a multi-scale gradient-matching term, whereas the other models both predict depth and estimate uncertainty. The present experiments do not isolate which of these design choices explain the performance difference. The comparison therefore highlights a trade-off between depth accuracy and uncertainty aware prediction rather than establishing an advantage of a particular loss formulation.

\begin{table*}[ht]
\centering
\begin{tabular}{llcccccccc}
\toprule
Method & Repr. &  AbsRel &  RMSE  &  AUSE  & AbsRel(M) & RMSE(M) & AbsRel(C) & RMSE(C)  \\
\midrule
Log-normal & 1 bin   & 0.2456         & 0.7259          &0.2294  & 0.2628          & 0.6899           & 0.1613          & 0.2080           \\
Log-normal & 5 bin   & 0.2015         & 0.6218          &       0.2609  & 0.2128          & 0.5693         & 0.1331          & 0.1938          \\
Log-normal & 10 bin  & 0.1914         & 0.5941          &   \textbf{0.1785}  & 0.2019          & 0.5538          & \textbf{0.0919} & \textbf{0.1226} \\
Log-normal & 20 bin  & 0.3333         & 0.7915          & 0.2443 & 0.3296          & 0.7195          & 0.1780          & 0.2364   \\
Log-normal & CSTR  & 0.2145         & 0.6633         &  0.2585    & 0.2237          & 0.6041          & 0.1542          & 0.2302  \\
Log-normal & TORE   & 0.2251         & 0.6429          &  0.2388  & 0.2582          & 0.6325         & 0.1295          & 0.1796         \\
Gaussian & 5 bin   & 0.2177         & 0.6664          &  0.2328   & 0.2403          & 0.6354          & 0.1500          & 0.2011          \\
Gaussian &  CSTR    & 0.2178         & 0.6466           &  0.2150     & 0.2410          & 0.6257          & 0.1214          & 0.1725          \\
EDL & 5 bin    & \textbf{0.1757} & \textbf{0.5788}  &  0.2368     & \textbf{0.1870} & \textbf{0.5370}  & 0.1117          & 0.1522          \\
EDL & CSTR    & 0.2776         & 0.8038           &  0.2183    & 0.2927          & 0.7457          & 0.1440          & 0.2156           \\
\midrule
E2Depth & 5 bin     & 0.1484         & 0.4667            & -     & 0.1698         & 0.4503         & - & -  \\
Log-normal & Grayscale  & 0.2752       & 0.7093           &  0.2984 & 0.3194        & 0.7260           & 0.2556          & 0.3218          \\
\bottomrule
\end{tabular}
\vspace{1cm}
\caption{Evaluation of the uncertainty models and event representations on the test sequence in terms of absolute relative error, root mean squared error, area under the sparsification error. (M) considers only pixels that generated events within the considered time window, approximately 10 \% of all pixels. (C) considers the pixels with lowest predicted uncertainty, selected to match the proportion of event-triggered pixels in (M). The best values (apart from E2Depth) for each metric are highlighted in bold. The test dataset is MVSEC indoor flying 1.}
\label{tab:my_depth_results}
\end{table*}

\begin{figure*}[htbp!]
  \centering
  \setlength{\tabcolsep}{1pt}
  \begin{tabular}{ccc}
    \includegraphics[width=0.25\textwidth]{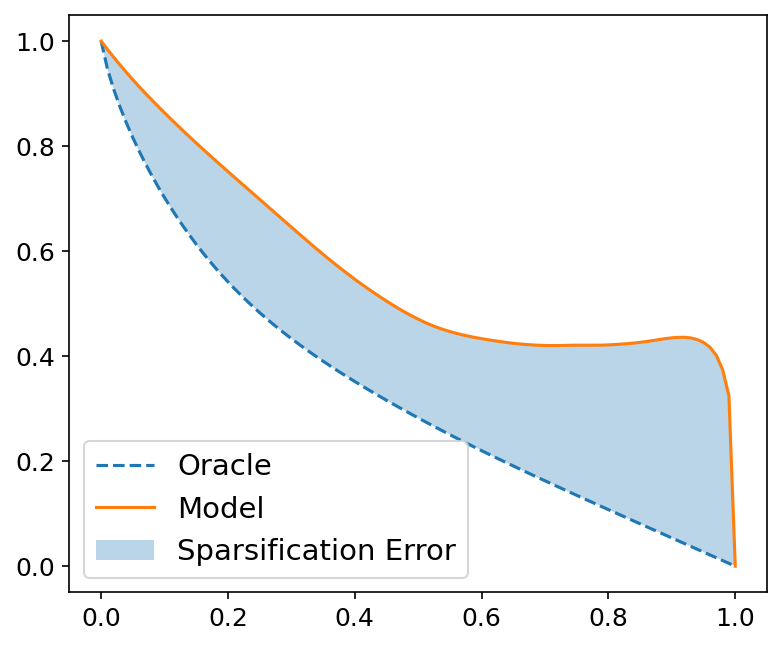} &
    \includegraphics[width=0.25\textwidth]{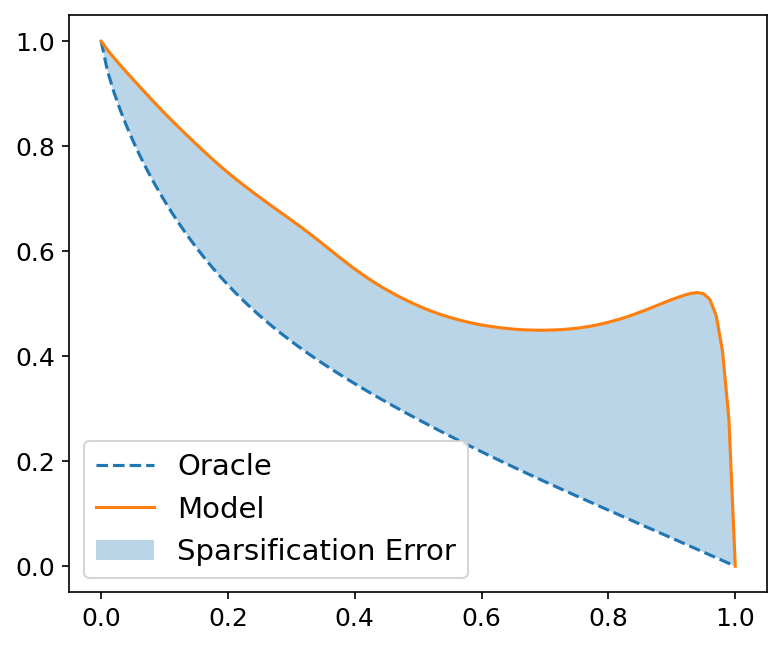} &
    \includegraphics[width=0.25\textwidth]{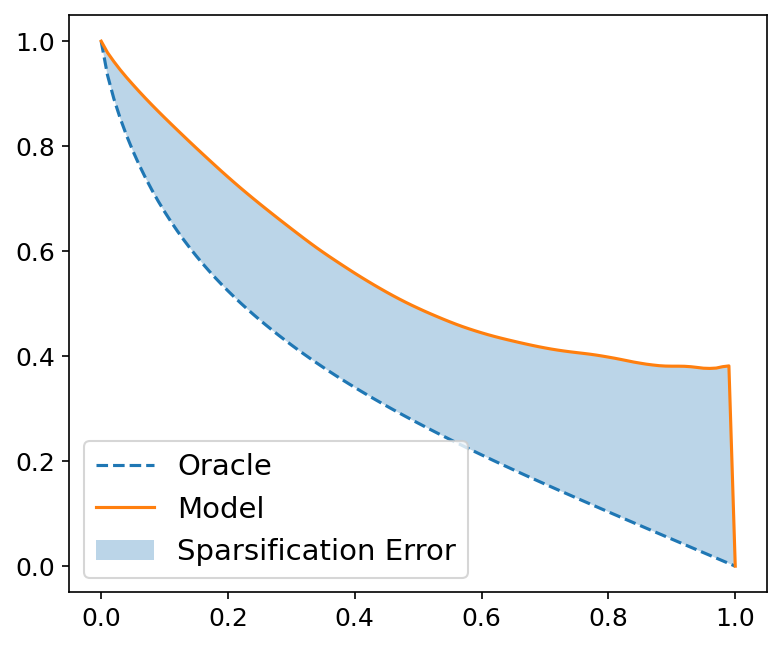} \\
     Gauss, 5 bin 
    &
    LogNorm, 5 bin 
    &
   EDL, 5 bin \\ 
  \end{tabular}
  \caption{Sparsification curves for some combinations of uncertainty models and event representations tested on MVSEC indoor flying 1.}
  \label{fig:all_graphs}
\end{figure*}

The sparsification curves for selected event representations are shown in \Cref{fig:all_graphs}. The Gaussian and log-normal 5-bin variants exhibit noticeable increases in error toward the right end of the curve (i.e., spikes as $\alpha \rightarrow 1$), indicating suboptimal ranking of highly uncertain predictions. In contrast, the EDL 5-bin configuration does not display this behavior, suggesting a more reliable ordering in high-uncertainty regions. Among all methods, the log-normal 10-bin model achieves the lowest AUSE, whereas the remaining configurations yield comparable AUSE values.

\begin{figure*}[t]
    \centering
    \setlength{\tabcolsep}{2pt}
    \newcommand{\imgw}{0.18\linewidth}

    \scriptsize
    \begin{tabular}{@{}c c c c@{}}
        \includegraphics[width=\imgw]{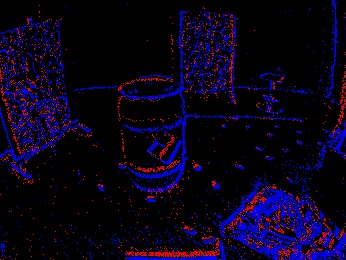} &
        \includegraphics[width=\imgw]{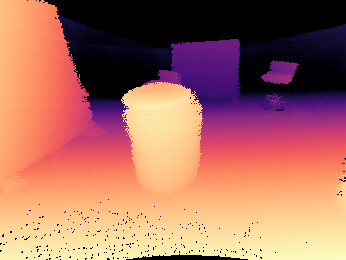} &
        \includegraphics[width=\imgw]{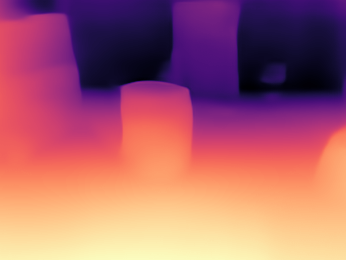} &
        \includegraphics[width=\imgw]{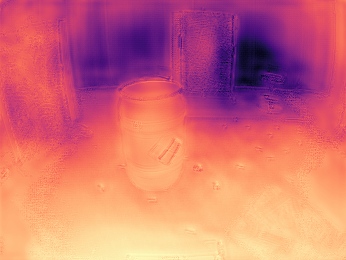} \\[1pt]
        \shortstack{Events} &
        \shortstack{Ground truth}  &
          \shortstack{E2Depth, 5 Bin} &
        \shortstack{LogNorm, Grayscale}
        \\[1pt]
        \includegraphics[width=\imgw]{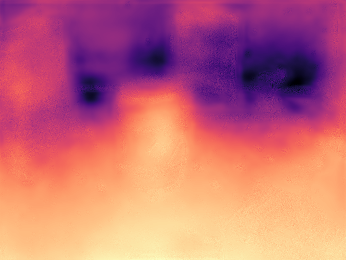} &
        \includegraphics[width=\imgw]{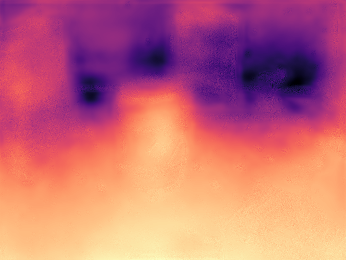} &
        \includegraphics[width=\imgw]{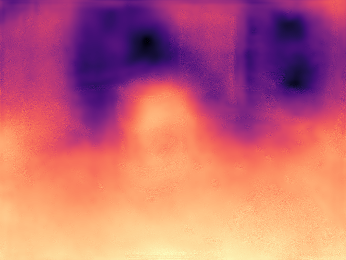} &
        \includegraphics[width=\imgw]{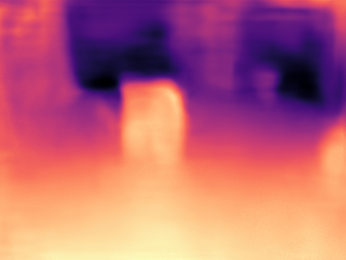} \\[1pt]
        \shortstack{LogNorm, CSTR} &
        Gauss, CSTR &
        \shortstack{LogNorm, 5 Bin} &
          \shortstack{EDL, 5 Bin} 
    \end{tabular}
    \caption{Comparison of depth predictions for six combinations of uncertainty modeling approaches and event representations. Orange indicates closer regions, while purple denotes greater depth. The predicted depth maps are visually similar across methods; however, E2Depth produces smoother results, likely due to the use of a multi-scale gradient matching loss.}
    \label{fig:mean_depth_representations}
\end{figure*}

Example depth predictions produced by the different models are shown in \Cref{fig:mean_depth_representations}. For this scene, the visual differences between our methods are subtle, which is consistent with the comparable quantitative performance observed across event representations. The predictions of E2Depth appear visually sharper, likely due to the use of a multi-scale gradient matching loss, which explicitly encourages the preservation of fine structural details.

\begin{figure*}[t]
    \centering
    \setlength{\tabcolsep}{2pt}
    \newcommand{\imgw}{0.18\linewidth} 

    \scriptsize
    \begin{tabular}{@{}c c c c c@{}}
        \includegraphics[width=\imgw]{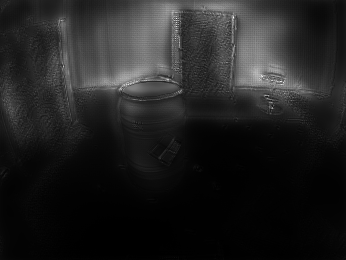} & 
        \includegraphics[width=\imgw]{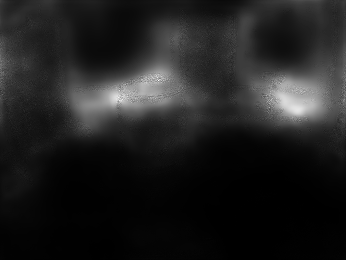}   &
       
        \includegraphics[width=\imgw]{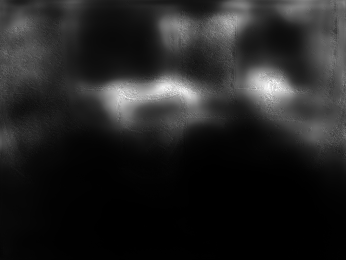} & 
        \includegraphics[width=\imgw]{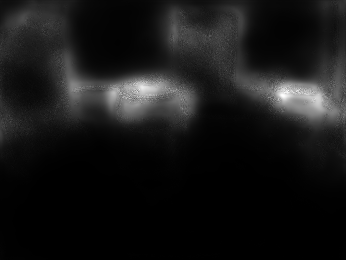} &
        \includegraphics[width=\imgw]{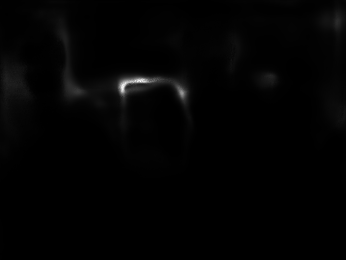} \\[1pt]

        \shortstack{LogNorm, Grayscale} &
         \shortstack{LogNorm, 5 Bin} &
        \shortstack{Gauss, CSTR} & 
        \shortstack{LogNorm, CSTR} &    
        \shortstack{EDL, 5 Bin}  \\
    \end{tabular}
    \caption{Comparison of predicted uncertainty across methods. Black indicates low predictive standard deviation (or epistemic uncertainty), while white corresponds to high uncertainty. The input events and ground truth depth are the same as those shown in \Cref{fig:mean_depth_representations}. The EDL 5-bin configuration exhibits overall lower predicted uncertainty, indicating higher confidence in depth predictions.}
  \label{fig:uncertainty_comp}
\end{figure*}

\Cref{fig:uncertainty_comp} compares predicted uncertainty maps across all methods. Black means low standard deviation, or epistemic error, and white means high. Ground truth is the same as in \Cref{fig:mean_depth_representations}.

\section{Discussion and future work}\label{sec:discussions_and_future_work}

The goal of this work has been to develop and evaluate monocular depth estimation networks that predict per-pixel depth together with uncertainty estimates from a single event camera stream. To this end, six event data representations were compared in combination with Gaussian, log-normal, and evidential deep learning (EDL) uncertainty modelling. The models were initially trained on synthetic BlinkVision sequences and subsequently fine-tuned on real data from the MVSEC dataset. Depth and uncertainty estimation performance were evaluated using absolute relative error (AbsRel), root mean squared error (RMSE), and the area under the sparsification error (AUSE).

In our experiments, the log-normal 10-bin and EDL 5-bin configurations achieved the best overall performance, although differences were generally modest. This suggests that the choice of event representation and uncertainty modelling has limited impact on depth accuracy within our experimental setting.

\Cref{tab:my_depth_results} further demonstrates that the predicted uncertainty successfully identifies pixels with low depth error. At a 10 \% retention rate, RMSE decreases by a factor of 4.8 and AbsRel by a factor of 2 in the log-normal 10-bin setting. Compared to the naive event-mask criterion M (which selects pixels that received events during the observation window), the uncertainty-based selection reduces AbsRel by approximately 1.5–2x across all tested variants. This indicates that the uncertainty estimates provide a reliable signal for identifying accurate predictions, largely independent of the specific uncertainty model or event representation.

Given the relatively small differences in raw accuracy across event representations, practical considerations such as inference speed, memory consumption, and suitability for real-time deployment may ultimately be the most decisive factors when selecting a representation for practical use. A systematic evaluation of these aspects is left for future work. From this perspective, it would also be valuable to explore additional event representations beyond those considered here. Promising alternatives include the Rasterized Event Point Cloud (RasEPC) and the Decoupled Event Voxel (DEV)~\cite{Yin_2023_CVIU}. Stacked TORE volumes also represent an interesting direction for further investigation~\cite{TORE-stereo}.

Finally, it can be noted that our predicted depth maps tend to be smooth, whereas actual depth is {\it piecewise smooth}, with sharp depth discontinuities. Earlier work has addressed this by incorporating a gradient matching loss term~\cite{Hidalgo20threedv}, which would probably be beneficial for our networks as well. Another way to improve the sharpness is to integrate self-attention into our encoder–decoder pipeline. This has proven beneficial in monocular depth estimation from RGB images~\cite{yang2021transformers}.


\bibliographystyle{IEEEtran}
\bibliography{thesis_refs.bib}

@Article{Hidalgo20threedv,
  author        = {Hidalgo-Carrio, Javier and Gehrig, Daniel and Scaramuzza, Davide},
  title         = {Learning Monocular Dense Depth from Events},
  journal       = {{IEEE} International Conference on 3D Vision.(3DV)},
  url           = {{http://rpg.ifi.uzh.ch/docs/3DV20_Hidalgo.pdf}},
  year          = 2020
}

@inproceedings{li2024blinkvisionbenchmarkopticalflow,
    title={BlinkVision: A Benchmark for Optical Flow, Scene Flow and Point Tracking Estimation using RGB Frames and Events},
    author={Yijin Li and Yichen Shen and Zhaoyang Huang and Shuo Chen, Weikang Bian and Xiaoyu Shi and Fu-Yun Wang and Keqiang Sun and Hujun Bao and Zhaopeng Cui and Guofeng Zhang and Hongsheng Li},
    booktitle={European Conference on Computer Vision (ECCV)},
    year={2024}
}

@inproceedings{DVS-Voltmeter,
author = {Lin, Songnan and Ma, Ye and Guo, Zhenhua and Wen, Bihan},
title = {DVS-Voltmeter: Stochastic Process-Based Event Simulator for Dynamic Vision Sensors},
year = {2022},
isbn = {978-3-031-20070-0},
url = {https://doi.org/10.1007/978-3-031-20071-7_34},
doi = {10.1007/978-3-031-20071-7_34},
booktitle = {Computer Vision – ECCV 2022},
}

@INPROCEEDINGS{zhu2018unsupervisedeventbasedlearningoptical,
  author={Alex Zihao Zhu and  Liangzhe Yuan and Kenneth Chaney and Kostas Daniilidis},
  booktitle={Conference on Computer Vision and Pattern Recognition (CVPR)}, 
  title={Unsupervised Event-Based Learning of Optical Flow, Depth and Egomotion}, 
  year={2019},
  }

@ARTICLE{MVSEC,
  author={Zhu, Alex Zihao and Thakur, Dinesh and Özaslan, Tolga and Pfrommer, Bernd and Kumar, Vijay and Daniilidis, Kostas},
  journal={IEEE Robotics and Automation Letters}, 
  title={The Multivehicle Stereo Event Camera Dataset: An Event Camera Dataset for 3D Perception}, 
  year={2018},
  volume={3},
  number={3},
  pages={2032-2039},
  doi={10.1109/LRA.2018.2800793}
}

@Article{RAMNet,
  author        = {Daniel Gehrig and Michelle Rüegg and Mathias Gehrig and Javier Hidalgo-Carrio and Davide Scaramuzza},
  title         = {Combining Events and Frames using Recurrent Asynchronous Multimodal Networks for Monocular Depth Prediction},
  journal       = {{IEEE} Robotic and Automation Letters. (RA-L)},
  url           = {http://rpg.ifi.uzh.ch/docs/RAL21_Gehrig.pdf},
  year          = 2021
}

@article{Xiliang,
title = {Uncertainty quantification metrics for deep regression},
journal = {Pattern Recognition Letters},
volume = {186},
pages = {91-97},
year = {2024},
issn = {0167-8655},
doi = {https://doi.org/10.1016/j.patrec.2024.09.011},
url = {https://www.sciencedirect.com/science/article/pii/S0167865524002733},
author = {Simon {Kristoffersson Lind} and Ziliang Xiong and Per-Erik Forssén and Volker Krüger}
}

@article{UNET,
  author = {Ronneberger, Olaf and Fischer, Philipp and Brox, Thomas},
  journal = {International Conference on Medical image computing},
  title = {U-Net: Convolutional Networks for Biomedical Image Segmentation.},
  year = 2015
}

@inproceedings{kingma2015adam,
  title     = {Adam: A Method for Stochastic Optimization},
  author    = {Kingma, Diederik P. and Ba, Jimmy},
  booktitle = {Proceedings of the 3rd International Conference on Learning Representations (ICLR)},
  year      = {2015},
  url       = {https://arxiv.org/abs/1412.6980}
}

@incollection{Event-Based-Stereo-Vision-with-Bio-Inspired-Silicon-Retina-Imagers,
author = {Jürgen Kogler and Christoph Sulzbachner and Martin Humenberger and Florian Eibensteiner},
title = {Address-Event Based Stereo Vision with Bio-Inspired Silicon Retina Imagers},
booktitle = {Advances in Theory and Applications of Stereo Vision},
publisher = {IntechOpen},
address = {Rijeka},
year = {2011},
editor = {Asim Bhatti},
chapter = {9},
doi = {10.5772/12941},
url = {https://doi.org/10.5772/12941}
}

@ARTICLE{Ontheuseoforientationfiltersfor3Dreconstructioninevent-drivenstereovision,
AUTHOR={Camunas-Mesa, Luis Alejandro  and Serrano-Gotarredona, Teresa  and Ieng, Sio Hoi  and Benosman, Ryad B. and Linares-Barranco, Bernabe },
TITLE={On the use of orientation filters for 3D reconstruction in event-driven stereo vision},
JOURNAL={Frontiers in Neuroscience},
VOLUME={Volume 8 - 2014},
YEAR={2014},
URL={https://www.frontiersin.org/journals/neuroscience/articles/10.3389/fnins.2014.00048},
DOI={10.3389/fnins.2014.00048},
ISSN={1662-453X}
}

@ARTICLE{AsynchronousEvent-BasedBinocularStereoMatching,
  author={Rogister, Paul and Benosman, Ryad and Ieng, Sio-Hoi and Lichtsteiner, Patrick and Delbruck, Tobi},
  journal={IEEE Transactions on Neural Networks and Learning Systems}, 
  title={Asynchronous Event-Based Binocular Stereo Matching}, 
  year={2012},
  volume={23},
  number={2},
  pages={347-353},
  doi={10.1109/TNNLS.2011.2180025}
}

@Article{Rebecq18ijcv,
  author        = {Henri Rebecq and Guillermo Gallego and Elias Mueggler and
                  Davide Scaramuzza},
  title         = {{EMVS}: Event-based Multi-View Stereo---{3D} Reconstruction
                  with an Event Camera in Real-Time},
  journal       = "Int. J. Comput. Vis.",
  year          = 2018,
  volume        = 126,
  issue         = 12,
  pages         = {1394--1414},
  month         = dec,
  doi           = {10.1007/s11263-017-1050-6}
}

@INPROCEEDINGS{CMapplications,
  author={Gallego, Guillermo and Rebecq, Henri and Scaramuzza, Davide},
  booktitle={2018 IEEE/CVF Conference on Computer Vision and Pattern Recognition}, 
  title={A Unifying Contrast Maximization Framework for Event Cameras, with Applications to Motion, Depth, and Optical Flow Estimation}, 
  year={2018},
  pages={3867-3876},
  doi={10.1109/CVPR.2018.00407}
}

@INPROCEEDINGS {CMfocus,
author = { Gallego, Guillermo and Gehrig, Mathias and Scaramuzza, Davide },
booktitle = { 2019 IEEE/CVF Conference on Computer Vision and Pattern Recognition (CVPR) },
title = {{ Focus Is All You Need: Loss Functions for Event-Based Vision }},
year = {2019},
doi = {10.1109/CVPR.2019.01256},
url = {https://doi.ieeecomputersociety.org/10.1109/CVPR.2019.01256},
publisher = {IEEE Computer Society},
}

@InProceedings{Shiba22eccv,
  author        = {Shintaro Shiba and Yoshimitsu Aoki and Guillermo Gallego},
  title         = {Secrets of Event-based Optical Flow},
  booktitle     = {European Conference on Computer Vision (ECCV)},
  pages         = {628--645},
  doi           = {10.1007/978-3-031-19797-0_36},
  year          = 2022
}

@INPROCEEDINGS {ActiveEventAlignment,
author = { Cai, Nan and Bideau, Pia },
booktitle = { 2025 IEEE/CVF Winter Conference on Applications of Computer Vision (WACV) },
title = {{ Active Event Alignment for Monocular Distance Estimation }},
year = {2025},
pages = {2464-2473},
doi = {10.1109/WACV61041.2025.00245},
publisher = {IEEE Computer Society},
}

@misc{meng2024learning,
  title        = {Learning Monocular Depth from Events via Egomotion Compensation},
  author       = {Meng, Haitao and Zhong, Chonghao and Tang, Sheng and JunJia, Lian and Lin, Wenwei and Bing, Zhenshan and Chang, Yi and Chen, Gang and Knoll, Alois},
  year         = {2024},
  eprint       = {2412.19067},
  archivePrefix= {arXiv},
  primaryClass = {cs.CV},
  note         = {arXiv preprint arXiv:2412.19067, submitted Dec.\ 26, 2024}
}

@inproceedings{DepthAnything,
  title={Depth Anything: Unleashing the Power of Large-Scale Unlabeled Data},
  author={Yang, Lihe and Kang, Bingyi and Huang, Zilong and Xu, Xiaogang and Feng, Jiashi and Zhao, Hengshuang},
  booktitle={CVPR},
  year={2024}
}

@article{cstr,
  title={CSTR: A Compact Spatio-Temporal Representation for Event-Based Vision},
  author={El Shair, Zaid A and Hassani, Ali and Rawashdeh, Samir A},
  journal={IEEE Access},
  year={2023},
  publisher={IEEE},
  pages={102899-102916},
  doi={10.1109/ACCESS.2023.3316143}
}

@ARTICLE{TORE,
  author={Baldwin, R. Wes and Liu, Ruixu and Almatrafi, Mohammed and Asari, Vijayan and Hirakawa, Keigo},
  journal={IEEE Transactions on Pattern Analysis and Machine Intelligence}, 
  title={Time-Ordered Recent Event ({TORE}) Volumes for Event Cameras}, 
  year={2023},
  volume={45},
  number={2},
  pages={2519-2532},
  doi={10.1109/TPAMI.2022.3172212}}

@inproceedings{TORE-stereo,
  title     = {TORE-Based Disparity Estimation In Stereo Event-Only Vision},
  author    = {Ruixu Liu and R. Wes Baldwin and Vijayan Asari and Keigo Hirakawa},
  booktitle = {Proceedings of the IEEE/CVF Conference on Computer Vision and Pattern Recognition (CVPR) Workshops},
  year      = {2021},
  note      = {CVPR 2021 Workshop on Event-Based Vision DSEC Competition Submission 1},
  url       = {https://dsec.ifi.uzh.ch/wp-content/uploads/sourcenova/uni-comp/cvprw-2021-competition/submissions/10/details.pdf}
}

@article{yin2024exploring,
  title   = {Exploring Event-based Human Pose Estimation with 3D Event Representations},
  author  = {Yin, Xiaoting and Shi, Hao and Chen, Jiaan and Wang, Ze and Ye, Yaozu and Yang, Kailun and Wang, Kaiwei},
  journal = {Computer Vision and Image Understanding},
  pages   = {104189},
  year    = {2024},
  publisher = {Elsevier},
  doi     = {10.1016/j.cviu.2024.104189}
}

@inproceedings{yang2021transformers,
  title={Transformer-Based Attention Networks for Continuous Pixel-Wise Prediction},
  author={Yang, Guanglei and Tang, Hao and Ding, Mingli and Sebe, Nicu and Ricci, Elisa},
  booktitle={ICCV},
  year={2021}
}

@article{Yin_2023_CVIU,
  title   = {Exploring Event-based Human Pose Estimation with 3D Event Representations},
  author  = {Yin, Xiaoting and Shi, Hao and Chen, Jiaan and Wang, Ze and Ye, Yaozu and Yang, Kailun and Wang, Kaiwei},
  journal = {Computer Vision and Image Understanding},
  year    = {2023},
}

@article{Event-survey,
author={Gallego,Guillermo and Delbruck,Tobi and Orchard,Garrick and Bartolozzi,Chiara and Taba,Brian and Censi,Andrea and Leutenegger,Stefan and Davison,Andrew J. and Conradt,Jorg and Daniilidis,Kostas and Scaramuzza,Davide},
year={2022},
title={Event-Based Vision: A Survey},
journal={IEEE transactions on pattern analysis and machine intelligence},
volume={44},
number={1},
pages={154-180},
isbn={0162-8828;1939-3539;},
language={English},
}

@inproceedings{zhou2018semi,
  title={Semi-dense 3D reconstruction with a stereo event camera},
  author={Zhou, Yi and Gallego, Guillermo and Rebecq, Henri and Kneip, Laurent and Li, Hongdong and Scaramuzza, Davide},
  booktitle={Proceedings of the European conference on computer vision (ECCV)},
  pages={235--251},
  year={2018}
}

@inproceedings{bartolomei2025depth,
  title={Depth AnyEvent: A Cross-Modal Distillation Paradigm for Event-Based Monocular Depth Estimation},
  author={Bartolomei, Luca and Mannocci, Enrico and Tosi, Fabio and Poggi, Matteo and Mattoccia, Stefano},
  booktitle={Proceedings of the IEEE/CVF International Conference on Computer Vision},
  pages={19669--19678},
  year={2025}
}

@article{sensoy2018evidential,
  title={Evidential deep learning to quantify classification uncertainty},
  author={Sensoy, Murat and Kaplan, Lance and Kandemir, Melih},
  journal={Advances in neural information processing systems},
  volume={31},
  year={2018}
}

@inproceedings{newcombe2011dtam,
  title={DTAM: Dense tracking and mapping in real-time},
  author={Newcombe, Richard A and Lovegrove, Steven J and Davison, Andrew J},
  booktitle={2011 international conference on computer vision},
  pages={2320--2327},
  year={2011},
  organization={IEEE}
}

@article{amini2020deep,
  title={Deep evidential regression},
  author={Amini, Alexander and Schwarting, Wilko and Soleimany, Ava and Rus, Daniela},
  journal={Advances in neural information processing systems},
  volume={33},
  pages={14927--14937},
  year={2020}
}

\end{document}